\documentstyle[aclap]{article} 
\input{psfig}
\title{Morphological
Disambiguation by Voting Constraints}
\author{Kemal Oflazer \and G\"{o}khan T\"{u}r \\
Department of Computer Engineering and Information Science \\
Bilkent University, Bilkent, TR-06533, Turkey\\
{\tt \{ko,tur\}@cs.bilkent.edu.tr}}
\begin{document}
\bibliographystyle{fullname}
\maketitle
\vspace{-0.5in}
\begin{abstract}
  We present a constraint-based morphological disambiguation system in
  which individual constraints vote on matching morphological parses,
  and disambiguation of all the tokens in a sentence is performed at
  the end by selecting parses that receive the highest votes.  This
  constraint application paradigm makes the outcome of the
  disambiguation {\em independent} of the rule sequence, and hence
  relieves the rule developer from worrying about potentially
  conflicting rule sequencing.  Our results for
  disambiguating Turkish indicate that using about 500 constraint
  rules and some additional simple statistics, we can attain a {\em
    recall of 95-96\%} and a {\em precision of 94-95\% with about 1.01
    parses per token.} Our system is implemented in Prolog and we are
  currently investigating an efficient implementation based on
  finite state transducers.
\end{abstract}

\section{Introduction}
\label{sec:intro}

Automatic morphological disambiguation is an important component in
higher level analysis of natural language text corpora.  There has
been a large number of studies in tagging and morphological
disambiguation using various techniques such as statistical techniques,
e.g., \cite{ChurchTagger,CuttingXeroxTagger,DeRoseCategoryOptimization}, constraint-based techniques
\cite{KarllsonConstraint,VoutilainenASyntax,VoutilainenHeikkila,VoutilainenTapanainen,OflazerANLP,OflazerEMNLP96}
and transformation-based techniques \cite{BrillSimpleRule,BrillAdvances,BrillCLArticle}.

This paper presents a novel approach to constraint based
morphological disambiguation which relieves the rule developer from
worrying about conflicting rule ordering requirements.
The approach depends on assigning votes to constraints according
to their complexity and specificity, and then letting  constraints
cast votes on matching parses of a given lexical item.  This approach does
not reflect the outcome of matching constraints to the set of
morphological parses immediately. Only after all applicable rules are applied to 
a sentence,  all tokens  are disambiguated in parallel.  
Thus, the outcome of the rule applications is independent of the 
order of rule applications.  Rule ordering issue has been discussed by
Voutilainen\shortcite{AtroThesis}, but he has recently
indicated\footnote{Voutilainen, Private communication.}
 that
insensitivity to rule ordering is not a property of their system
(although Voutilainen\shortcite{VoutilainenInConstraint} states that it is
a very desirable property) but
rather is achieved by extensively testing and tuning the rules.

In the following sections, we present an overview of the morphological
disambiguation problem, highlighted with examples from Turkish.  We
then present our approach and results.  We finally conclude with a
very brief outline of our investigation into efficient implementations
of our approach.

\section{Morphological Disambiguation}

In all languages, words are usually ambiguous in their parts-of-speech
or other morphological features, and may represent lexical items of
different syntactic categories, or morphological structures depending
on the syntactic and semantic context. In languages like English,
there are a very small number of possible word forms that can be
generated from  a given root
word, and a small number of part-of-speech tags associated with a given lexical form.
On the other hand, in languages like Turkish or Finnish with very
productive agglutinative morphology, it is possible to produce
thousands of forms (or even millions
\cite{HankamerMorphologicalParsing}) from a given root word and the
kinds of ambiguities one observes are quite different than what is
observed in languages like English.

In Turkish, there are ambiguities of the sort typically found in
languages like English (e.g., book/noun vs book/verb type). However, the
agglutinative nature of the language usually helps resolution of such
ambiguities due to the restrictions on morphotactics of subsequent morphemes. On the other
hand, this very nature introduces another kind of ambiguity, where a
lexical form can be morphologically interpreted in many ways not
usually predictable in advance.  Furthermore, Turkish allows very
productive derivational processes and the information about the
derivational structure of a word form is usually crucial for
disambiguation \cite{OflazerEMNLP96}.

Most kinds of morphological ambiguities that we have observed in
Turkish typically fall into one the following classes:\footnote{Output
  of the morphological analyzer is edited for clarity, and English
  glosses have been given. We have also provided the morpheme
  structure, where [\ldots]s, indicate elision. Glosses are given as
  linear feature value sequences corresponding to the morphemes (which
  are not shown). The feature names are as follows: {\tt CAT}-major
  category, {\tt TYPE}-minor category, {\tt ROOT}-main root form, {\tt
    AGR} -number and person agreement, {\tt POSS} - possessive
  agreement, {\tt CASE} - surface case, {\tt CONV} - conversion to the
  category following with a certain suffix indicated by the argument
  after that, {\tt TAM1}-tense, aspect, mood marker 1, {\tt
    SENSE}-verbal polarity. Upper cases in morphological output
  indicates one of the non-ASCII special Turkish characters: e.g.,
  {\tt G} denotes \u{g}, {\tt U} denotes \"{u}, etc.}

\begin{enumerate}
\item the form is uninflected and assumes the default inflectional
      features, e.g., 
\begin{small}
\begin{verbatim}
1. taS    (made of stone)
   [[CAT=ADJ][ROOT=taS]]
2. taS    (stone)
   [[CAT=NOUN][ROOT=taS]
    [AGR=3SG][POSS=NONE][CASE=NOM]]
3. taS    (overflow!)
   [[CAT=VERB][ROOT=taS][SENSE=POS]
    [TAM1=IMP][AGR=2SG]]
\end{verbatim}
\end{small}
\item Lexically different affixes (conveying 
  different morphological features) surface
  the same due to the morphographemic context, e.g.,
\begin{small}
\begin{verbatim}
1. ev+[n]in  (of the house)
   [[CAT=NOUN][ROOT=ev]
    [AGR=3SG][POSS=NONE][CASE=GEN]]
2. ev+in       (your house)
   [[CAT=NOUN][ROOT=ev]
    [AGR=3SG][POSS=2SG][CASE=NOM]]
\end{verbatim}    
\end{small}
\item The root of one of the parses is a prefix string  of the root of 
the
  other parse, and the parse with the shorter root word has a suffix
  which surfaces as the rest of the longer root word, e.g., 
\begin{small}
\begin{verbatim}
1. koyu+[u]n  (your dark (thing))
   [[CAT=ADJ][ROOT=koyu][CONV=NOUN=NONE]
    [AGR=3SG][POSS=2SG][CASE=NOM]]
2. koyun   (sheep)
   [[CAT=NOUN][ROOT=koyun]
    [AGR=3SG][POSS=NONE][CASE=NOM]]
3. koy+[n]un  (of the bay)
   [[CAT=NOUN][ROOT=koy]
    [AGR=3SG][POSS=NONE][CASE=GEN]]
4. koy+un     (your bay)
   [[CAT=NOUN][ROOT=koy]
    [AGR=3SG][POSS=2SG][CASE=NOM]]
5. koy+[y]un  (put!)
   [[CAT=VERB][ROOT=koy][SENSE=POS]
    [TAM1=IMP][AGR=2PL]]
\end{verbatim}
\end{small}
\item The roots take different numbers of unrelated inflectional
  and/or derivational suffixes which when concatenated turn out to
  have the same surface form, e.g.,
\begin{small}
\begin{verbatim}
1.  yap+madan   (without having done (it)) 
   [[CAT=VERB][ROOT=yap][SENSE=POS]
    [CONV=ADVERB=MADAN]]
2.  yap+ma+dan  (from doing (it))
   [[CAT=VERB][ROOT=yap][SENSE=POS]
    [CONV=NOUN=MA][TYPE=INFINITIVE]
    [AGR=3SG][POSS=NONE][CASE=ABL]]
\end{verbatim}
\end{small}
  
\item One of the ambiguous parses is a lexicalized form while
  another is  form derived by a productive derivation as in 1
  and 2 below.

\item The same suffix appears in different positions in the morphotactic
  paradigm conveying different information as in 2 and 3 below.
\begin{small}
\begin{verbatim}
1. uygulama /   (application)
   [[CAT=NOUN][ROOT=uygulama]
    [AGR=3SG][POSS=NONE][CASE=NOM]]
2. uygula+ma /  ((the act of) applying)
    [[CAT=VERB][ROOT=uygula][SENSE=POS]
     [CONV=NOUN=MA][TYPE=INFINITIVE]
     [AGR=3SG][POSS=NONE][CASE=NOM]]
3. uygula+ma /  (do not apply!)
    [[CAT=VERB][ROOT=uygula][SENSE=NEG]
     [TAM1=IMP][AGR=2SG]]
\end{verbatim}
\end{small}
\end{enumerate}

The main intent of our system is to achieve morphological
disambiguation by choosing for a given ambiguous token, the
correct parse in a given context. It is certainly possible that a
given token may have multiple correct parses, usually with the same
inflectional features, or with inflectional features not ruled out by
the syntactic context, but one will be the ``correct'' parse usually 
on semantic grounds.

We consider a token {\em fully disambiguated} if it has only one
morphological parse remaining after automatic disambiguation. We
consider a token as correctly disambiguated, if one of the parses
remaining for that token is the {\em correct} intended
parse.
 We evaluate the resulting disambiguated text by a number of metrics
defined as follows \cite{VoutilainenInConstraint}:
\[ 
 Ambiguity = \frac{\# Parses}{\# Tokens }
\]

\[
Recall =  \frac{ \# Tokens\  Correctly\  Disambiguated}
                    {\# Tokens}
\]

\[
Precision = \frac{\# Tokens\  Correctly\  Disambiguated}
                    {\# Parses}
 \]
 In the ideal case where each token is uniquely and correctly
 disambiguated with the correct parse, both recall and precision will
 be 1.0. On the other hand, a text where each token is annotated with
 all possible parses,\footnote{Assuming no unknown words.} the recall
 will be 1.0, but the precision will be low. The goal is to have both
 recall and precision as high as possible.

\section{Constraint-based  Morphological Disambiguation}
This section outlines our  approach to constraint-based morphological
disambiguation where constraints vote on matching parses of 
sequential tokens.

\subsection{ Constraints on morphological parses}

We describe constraints on the morphological parses of tokens using 
rules with two components 
\begin{quote}
 $ R= (C_{1},C_{2},\cdots,C_{n}; V)$
\end{quote}
where the C$_{i}$  are (possibly hierarchical)  feature constraints on 
a sequence of 
the morphological parses, and $V$ is an integer denoting 
the vote of the rule.

To illustrate the flavor of our rules we can give the following
examples:
\begin{enumerate}
\item  The  following  rule with two constraints matches parses with case
feature ablative, preceding a parse matching a postposition
subcategorizing for an ablative nominal form.
  \begin{small}
    \begin{verbatim}
 [[case:abl],[cat:postp,subcat:abl]]
\end{verbatim}
  \end{small}
  \item  The rule 
  \begin{small}
\begin{verbatim}
[[agr:'2SG',case:gen],[cat:noun,poss:'2SG']]
\end{verbatim}
  \end{small}
matches a nominal form with a possessive marker {\tt 2SG},
following a pronoun with {\tt 2SG} agreement and
genitive case, enforcing the simplest form of  noun
phrase constraints.   

\item In general  constraints can
make references to the derivational structure of the lexical form and
hence be hierarchical.  
For instance, the following rule is an example
of a rule employing a hierarchical constraint:
\begin{small}
\begin{verbatim}
[[cat:adj,stem:[tam1:narr]],
          [cat:noun,stem:no]]
\end{verbatim}
\end{small}
which matches the derived participle reading of a verb with 
narrative
past tense, if it is followed by an underived noun parse.
\end{enumerate}

\subsection{Determining the vote of a rule}
There are a number of ways votes can be assigned to rules. For the
purposes of this work the vote of a rule is determined by its static
properties, but it is certainly conceivable that votes can be assigned
or learned by using statistics from disambiguated corpora.\footnote{We
  have left this for future work.} For static vote assignment,
intuitively, we would like to give high votes to rules that are more
specific: i.e., to rules that have
\begin{itemize}
\item higher number of constraints,
\item higher number of features in the constraints,
\item constraints that make reference to  nested stems (from which the
  current form is derived),
\item constraints that make reference to  very specific features or values.
\end{itemize}
Let  $ R= (C_{1},C_{2},\cdots,C_{n}; V)$ be a constraint rule. The 
vote
$V$ is determined as
\[ V=\sum_{i=1}^{n}V(C_i)\]
where $V(C_i)$ is the contribution of constraint $C_i$ to the vote of
the rule  $R$.  A (generic) constraint has the following form:
\[C=[(f_1 : v_1) \& (f_2 : v_2) \& \cdots (f_m : v_m)] \]
where $f_i$  is the name of a morphological feature, and $v_i$ is one of the
possible values for that feature. The contribution  of
$f_i :  v_i$  in the vote
of a constraint depends on a number of factors:
\begin{enumerate}
\item The value $v_i$ may be a distinguished value that has a more
  important function in dis\-am\-bi\-gu\-a\-tion.\footnote{For instance, for
Turkish we have no\-ted that the ge\-ni\-ti\-ve case marker is usually very
helpful in disambiguation.} In this case, the weight of the
  feature constraint  is $w(v_i) (> 1)$. 
\item The  feature itself may be  a distinguished feature
  which has more important function in disambiguation. In this case the
  weight of the feature is $w(f_i) (>1)$. 
\item If the feature $f_i$ refers to the stem of a derived form  and
  the value part of the feature constraint is a full fledged
  constraint $C'$ on the stem structure,  the weight of 
the
  feature constraint is found by recursively computing the vote of
   $C'$ and  scaling the resulting value by a factor (2 in our current
   system) to improve its specificity.
\item Otherwise, the weight of the feature constraint is 1.
\end{enumerate}
For example suppose we have the following  constraint:
\begin{small}
\begin{verbatim}
[cat:noun, case:gen,
    stem:[cat:adj, stem:[cat:v], suffix=mis]]
\end{verbatim}
\end{small}
Assuming the value {\tt gen } is a distinguished value with weight 4 
(cf., factor 1 above),
the vote of this constraint is computed as follows: 
\begin{enumerate}
\item {\tt cat:noun} contributes 1,
\item {\tt case:gen} contributes 4,
\item {\tt stem:[cat:adj, stem:[cat:v],suffix=mis]} contributes
8 computed as follows:
\begin{enumerate}
\item {\tt cat:adj} contributes 1,
\item {\tt suffix=mis} contributes 1,
\item {\tt stem:[cat:v]} contributes $2 = 2 * 1$, the 1 being from 
{\tt
    cat:v},
\item the sum 4 is scaled by 2 to give 8.
\end{enumerate}
\item Votes from steps 1, 2 and 3(d) are added up to give 13 as the constraint
vote.
\end{enumerate}

We also employ a set of rules which express preferences among the
parses of single lexical form independent of the context in which the
form occurs.  The weights for these rules are currently manually 
determined. These rules give negative votes to the parses which are
not preferred or high votes to certain parses which are always preferred.
Our experience is that such preference rules depend on
the kind of the text one is disambiguating. For instance if one is
disambiguating a manual of some sort, imperative readings of verbs are
certainly possible, whereas in normal plain text with no discourse,
such readings are discouraged. 

\subsection{Voting and selecting parses}
A rule $ R= (C_{1},C_{2},\cdots,C_{n}; V)$ will match
a sequence of tokens $w_{i},w_{i+1}, \cdots, w_{i+n-1}$ within a
sentence $w_1$ through $w_s$ if {\em some morphological parse of 
every token} $w_{j}, i
\leq j \leq i+n-1$ is subsumed by the corresponding constraint {\tt
  C$_{j-i+1}$}. When all constraints match, the votes of all the
matching parses  are incremented by $V$.  If a given constraint
matches more than one parse of a token, then the votes of all such matching
parses are incremented.

After all rules have been applied to all token positions in a
sentence and votes are tallied, morphological parses are selected in
the following manner. Let $v_l$ and $v_h$ be the votes of the lowest
and highest scoring parses for a given token. All parses with votes
equal to or higher than $v_l + m * (v_h - v_l)$ are selected with 
$m \   
 (0 \leq m \leq 1)$  being a parameter. $m = 1$ selects the highest
 scoring parse(s).

\section{Results from Disambiguating Turkish Text}
We have applied our approach to disambiguating Turkish text.  Raw text
is processed by a {\em preprocessor} which  segments the text  into sentences 
using various heuristics about 
punctuation, and then tokenizes and runs it through a wide-coverage
high-performance morphological analyzer developed using two-level
morphology tools by Xerox \cite{KarttunenLexc}.  The preprocessor module
also performs a number of additional functions such as grouping of
{\em lexicalized} and {\em non-lexicalized} collocations, compound
verbs, etc., \cite{OflazerANLP,OflazerEMNLP96}.
The preprocessor also uses a second morphological processor for
dealing with unknown words which recovers any derivational and
inflectional information from a word {\em even if the root word is not
known.}  This unknown word processor has a (nominal) root lexicon which
recognizes $S^+$, where $S$ is the Turkish surface alphabet (in the
two-level morphology sense), but then tries to interpret an arbitrary
postfix string of the unknown word, as a sequence of Turkish suffixes 
subject to all morphographemic constraints \cite{OflazerEMNLP96}.  

We have applied our approach to four texts labeled ARK, HIST, MAN,
EMB, with statistics given in Table \ref{tab:textstat}.  The tokens
considered are those that are generated after morphological analysis,
unknown word processing and any lexical coalescing is done.  The words
that are counted as unknown are those that could not even be processed
by the unknown noun processor as they violate Turkish morphographemic
constraints.   Whenever an unknown word has more than
one parse it is counted under the appropriate group.\footnote{The
  reason for the (comparatively) high number of unknown words in MAN,
  is that tokens found in such texts, like {\em f10}, denoting a
  function key in the computer can not be parsed as a Turkish root
  word!} The fourth and fifth columns in this table give the average
parses per token and the initial precision assuming initial recall is
100\%.

{\small
\begin{table*}[htbp]
  \begin{center}
    \begin{tabular}[c]{|l|r|r|r|r|r|r|r|r|r|r|}
      \hline 
     & & & & & \multicolumn{6}{c|}{Distribution} \\
     & & & & & \multicolumn{6}{c|}{of} \\
    \multicolumn{1}{|c|}{Text}  & \multicolumn{1}{c|}{Sent.} & 
\multicolumn{1}{c|}{Tokens} & \multicolumn{1}{c|}{Parses/ }& 
\multicolumn{1}{c|}{Init. }&\multicolumn{6}{c|}{Morphological
      Parses} \\ \cline{6-11} 
    & & & \multicolumn{1}{c|}{Token }& \multicolumn{1}{c|}{Prec. }& 0 & 1 & 2 & 3 & 4 & $> 4$ \\ \hline
ARK & 492  & 7928 & 1.823 &     0.55 & 0.15\% & 49.34\% & 30.93\% & 9.19\% & 8.46\%& 
1.93\%  \\ \hline
HIST  & 270 & 5212 & 1.797& 0.56& 0.02\% & 50.63\% & 30.68\% & 8.62\% & 8.36\% &
1.69\% \\ \hline
MAN   & 204 & 2756 & 1.840&     0.54& 0.65\% & 49.01\% & 31.70\% & 6.37\% & 8.91\% & 
3.36\%\\ \hline
EMB & 198 & 5177& 1.914 &       0.52& 0.09\% & 43.94\% & 34.58\% & 9.60\% & 9.46\% & 
2.33\%\\ \hline
    \end{tabular}
  \end{center}
  \caption{Statistics on Texts}
  \label{tab:textstat}
\end{table*}
}

We have disambiguated these texts using a rule base of about 500
hand-crafted rules. Most of the rule crafting was done using the
general linguistic constraints and constraints that we derived from
the first text, ARK. In this sense, this text is our ``training
data'', while the other three texts were not considered in rule
crafting.

Our results are summarized in Table \ref{Table:VoteResult}. The last
four columns in this table present results for different values for
the parameter $m$ mentioned above, $m = 1$ denoting the case when only
the highest scoring parse(s) is (are) selected. The columns for $m <
1$ are presented in order to emphasize that drastic loss  of
precision for those cases. Even at $m=0.95$ there is considerable loss
of precision and going up to $m=1$ causes a dramatic increase in
precision without a significant loss in recall. It can be seen that we
can attain very good recall and quite acceptable precision with just
voting constraint rules.  Our experience is that we can in principle
add highly specialized rules by covering a larger text base to improve
our recall and precision for the $m=1$.  A post-mortem analysis has
shown that cases that have been missed are mostly due to
morphosyntactic dependencies that span a context much wider that 5
tokens that we currently employ.

\begin{table}[htbp]
  \begin{center}
    \begin{tabular}{{|l|l|r||r|r|r|}}
\hline
\multicolumn{2}{|c|}{}  & \multicolumn{4}{c|}{Vote Range Selected(m)} 
\\ \hline
\multicolumn{2}{|c|}{TEXT} & 1.0 & 0.95 &0.8 &0.6 \\ \hline
ARK & Rec. & 98.05 & 98.47 & 98.69 & 98.77\\ \hline
& Prec. & 94.13& 87.65 & 84.41 & 82.43 \\ \hline
& Amb. & 1.042& 1.123& 1.169& 1.200 \\ \hline
\hline
HIST & Rec. & 97.03& 97.65 &98.81 & 97.01 \\ \hline
& Prec. & 94.13&87.10&84.41 & 82.29\\ \hline
& Amb. &1.058 &1.121 & 1.169 & 1.189\\ \hline
\hline
MAN & Rec. & 97.03 & 97.92 & 97.81 & 98.77\\ \hline
& Prec. &91.05 & 83.51 &  79.85& 77.34\\ \hline
& Amb. &1.068 &  1.172 & 1.237 & 1.277 \\ \hline
\hline
EMB & Rec. & 96.51 & 97.48  &97.76 &97.94\\ \hline
& Prec. & 91.28& 84.36 & 77.87 & 75.79\\ \hline
& Amb. & 1.057 &  1.150 & 1.255& 1.292\\ \hline
    \end{tabular}
  \end{center}
  \caption{Results with voting constraints}
  \label{Table:VoteResult}
\end{table}

\subsection{Using  root and contextual statistics}
We have employed two additional sources of information: {\em root word
usage statistics, and contextual statistics. }
We have statistics  compiled from previously disambiguated text, on root
frequencies. After the application of constraints as described above,
for tokens which are still ambiguous with  ambiguity  resulting
from different root words, we discard  parses if the frequencies
of the root words for those parses are  considerably lower  than the 
frequency of the root of the highest scoring parse.  The results after 
applying this step on top of voting, with $m=1$, are shown in the fourth column of Table
\ref{Table:VoteResult2} (labeled V+R).  

\begin{table}[htbp]
  \begin{center}
    \begin{tabular}{|l|l|r|r|r|}
\hline
\multicolumn{2}{|c|}{TEXT} & \multicolumn{1}{c|}{V} & \multicolumn{1}{c|}{V+R} & \multicolumn{1}{c|}{V+R+C} \\ \hline
 ARK & Rec. &   98.05 & 97.60 &  96.98 \\ \hline
 & Prec. &   94.13 & 95.28 &  96.19\\ \hline
 & Amb. &  1.042 & 1.024 &  1.008\\ \hline
\hline
 HIST & Rec. &   97.03 & 96.52 &   95.62 \\ \hline
 & Prec. &   94.13 & 92.59 &  94.33\\ \hline
 & Amb. &  1.058 & 1.042 &  1.013 \\ \hline
\hline
 MAN & Rec. &   97.03 & 96.47 &  95.84 \\ \hline
 & Prec. & 91.05 & 93.08 &  94.47 \\ \hline
 & Amb. &1.058 & 1.042 &  1.014 \\ \hline
\hline
EMB & Rec. &   96.51 & 96.47 &  95.37\\ \hline
& Prec. &   91.28 & 93.08 &  94.45\\ \hline
& Amb. &   1.057 & 1.036 &  1.009 \\ \hline
    \end{tabular}
  \end{center}
  \caption{Results with voting constraints and
    root statistics, context statistics}            %

  \label{Table:VoteResult2}
\end{table}

On top of this, we use the following heuristic using context
statistics to eliminate any further ambiguities.  For every remaining
ambiguous token with unambiguous immediate left and right contexts
(i.e., the tokens in the immediate left and right are unambiguous), we
perform the following, {\em by ignoring the root/stem feature of the
  parses}:
\begin{enumerate}
\item  For every ambiguous parse in such an unambiguous context,  we 
count how many times,  this parse occurs {\em unambiguously} in 
exactly the same unambiguous context, in the rest of the text.

\item We then choose the parse whose count is substantially higher 
than the others.
\end{enumerate}
The results after 
applying this step on of the previous two steps  are shown in the last column of  Table
\ref{Table:VoteResult2} (labeled V+R+C). One can see from the last three columns of
this table, the impact of each of the steps.

By ignoring root/stem features during this process, we essentially 
are considering just the top level inflectional information of the 
parses. This is very similar to Brill's use of contexts to induce 
transformation rules for his tagger 
\cite{BrillSimpleRule,BrillCLArticle}, but instead of generating 
transformation rules from a training text, we gather statistics and 
apply them to parses in the text being disambiguated.

\section{Efficient Implementation Techniques  and Extensions}

\begin{figure*}
\centerline{\psfig{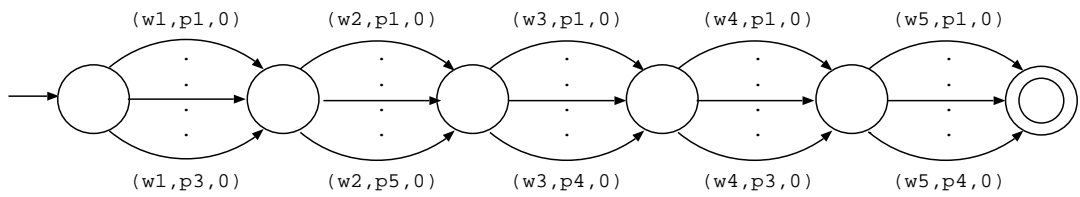}}
\caption{Sentence as a finite state recognizer.}
\label{fig:FSA}
\end{figure*}

The current implementation of the voting approach is meant to be a
proof of concept implementation and is rather inefficient.  However,
the use of regular relations and finite state transducers
\cite{KaplanKayCLArticle} provide a very efficient implementation
method.  For this, we view the parses of the tokens making up a
sentence as making up a acyclic a finite state recognizer with the
states marking word boundaries and the ambiguous interpretations of
the tokens as the state transitions between states, the rightmost node
denoting the final state, as depicted in Figure \ref{fig:FSA} for a
sentence with 5 tokens.  In Figure \ref{fig:FSA}, the transition
labels are triples of the sort $(w_i, p_j, 0)$ for the $j^{th}$ parse
of token $i$, with the 0 indicating the initial vote of the parse.
The rules imposing constraints  can also be  represented
as transducers which increment the votes of the matching transition
labels by an appropriate amount.\footnote{Suggested by Lauri Karttunen
  (private com\-mu\-ni\-ca\-tion).} Such transducers ignore  and pass
through unchanged, parses  that they are not sensitive to.

\begin{figure*}
\centerline{\psfig{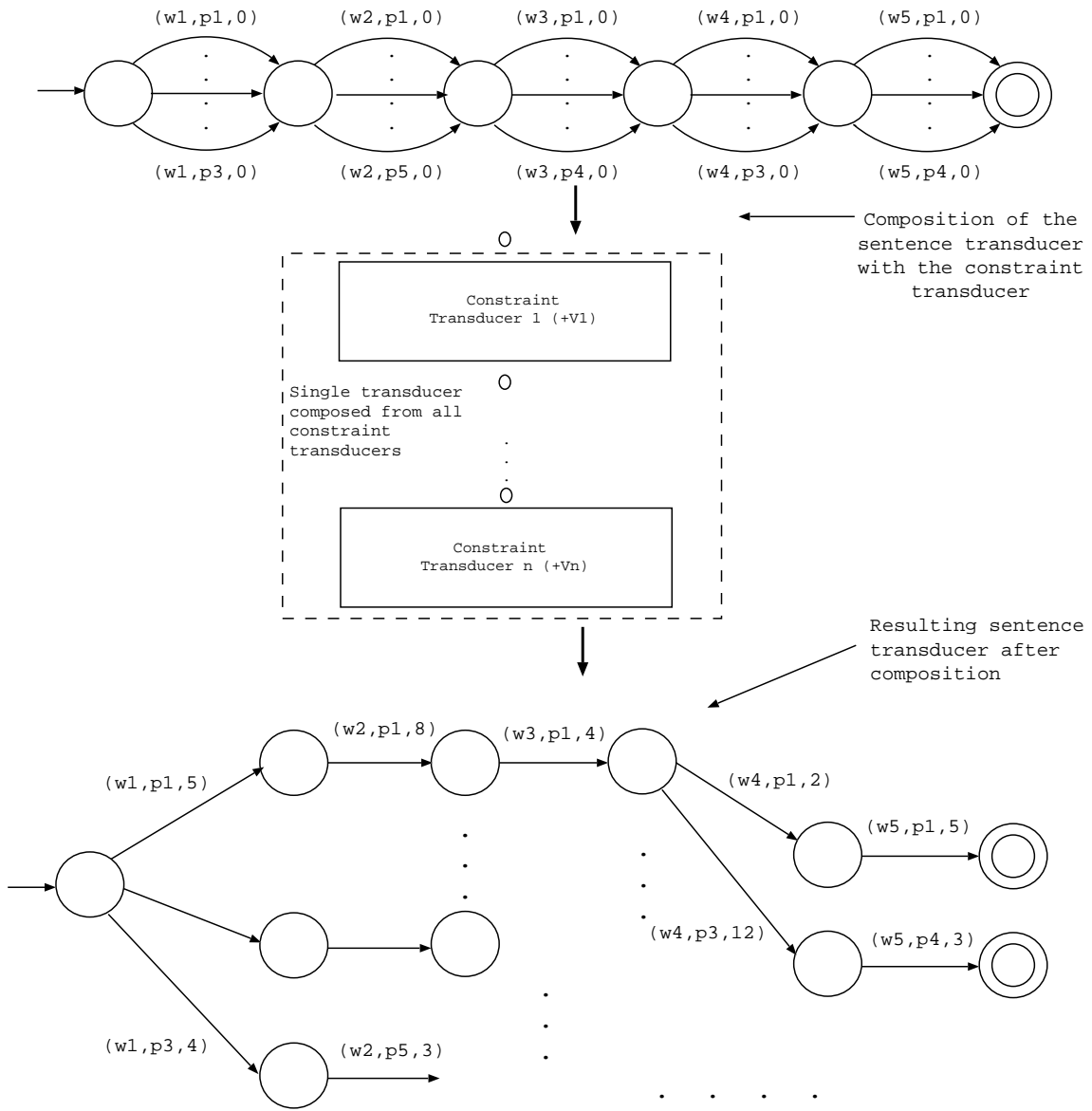}}
\caption{Sentence and Constraint Transducers}
\label{fig:compose}
\end{figure*}

When a finite state recognizer corresponding to the input sentence
(which actually may be considered as an identity transducer) is
composed with a constraint transducer, one gets a slightly modified
version of the sentence transducer with possibly additional
transitions and states, where the votes of some of the labels have
been appropriately incremented.  When the sentence transducer is
composed with all the constraint transducers in sequence, all possible votes are
cast and the final sentence transducer reflects all the votes. The
parse corresponding to each token with the highest vote can then be
selected. {\em The key point here is that due to the nature of the
composition operator, the constraint transducers can be composed
off-line first, giving a single constraint transducer and then this
one is composed with every sentence transducer once} (See Figure
\ref{fig:compose}).

The idea of voting can further be extended to a {\em path voting}
framework where rules vote on paths containing sequences of matching
parses and the path from the start state to the final stated with the
highest votes received, is then selected. This can be implemented
again using finite state transducers as described above (except that
path vote is apportioned equally to relevant parse votes), but instead
of selecting highest scoring parses, one selects the path from the
start state to one of the final states where the sum of the parse
votes is maximum.  We have recently completed a prototype implementation of
this approach (in C) for English (Brown Corpus) and have obtained
quite similar  results \cite{TurEMNLP97TR}.

\section{Conclusions}
We have presented an approach to constraint-based morphological
disambiguation which uses constraint voting as its primary mechanism
for parse selection and alleviates the rule developer from worrying
about rule ordering issues. Our approach is quite general and is
applicable to any language.  Rules describing language specific
linguistic constraints vote on matching parses of tokens, and at the
end, parses for every token receiving the highest tokens are selected.
We have applied this approach to Turkish, a language with complex
agglutinative word forms exhibiting morphological ambiguity phenomena
not  usually found in languages like English and have obtained quite promising
results. The convenience of adding new rules in without worrying about
where exactly it goes in terms of rule ordering (something that
hampered our progress in our earlier work on disambiguating Turkish
morphology \cite{OflazerANLP,OflazerEMNLP96}), has also been a key
positive point.  Furthermore, it is also possible to use rules with
negative votes to disallow impossible cases. This has been quite
useful for our work on tagging English \cite{TurEMNLP97TR} where such
rules with negative weights were used to fine tune the behavior of
the tagger in various problematic cases. 

The proposed approach is also amenable to an efficient implementation
by finite state transducers \cite{KaplanKayCLArticle}. By using
finite state transducers, it is furthermore possible to 
use a bit more expressive rule formalism including for instance the
Kleene * operator so that one can use a much smaller set of rules to
cover the same set of local linguistic phenomena.

Our current and future work in this framework involves  the learning of
constraints and their votes from corpora, and combining  learned and
hand-crafted rules.

\section{Acknowledgments}
This research has been supported in part by a NATO Science for
Stability Grant TU--LANGUAGE. We thank Lauri Karttunen of Rank Xerox
Research Centre in Grenoble for providing the Xerox two-level
morphology tools on which the Turkish morphological analyzer was
built.

\bibliography{tagging}

\end{document}